\documentclass[runningheads]{llncs}

\usepackage[year=2026]{eccv}

\usepackage{eccvabbrv}

\usepackage[numbers]{natbib}

\usepackage{graphicx}
\usepackage{booktabs}
\usepackage{wrapfig}

\usepackage[accsupp]{axessibility}
\definecolor{cvprblue}{rgb}{0.21,0.49,0.74}
\usepackage[pagebackref,breaklinks,colorlinks,allcolors=cvprblue]{hyperref}
\usepackage{booktabs}
\usepackage[table]{xcolor}
\usepackage{algorithm}
\usepackage{algorithmicx}
\usepackage{algpseudocode}
\usepackage{amsmath}
\usepackage{hhline}

\usepackage{booktabs}
\usepackage[table]{xcolor}
\usepackage{colortbl}
\usepackage{makecell}
\usepackage{multirow}
\usepackage{soul}

\usepackage{tikz}
\usetikzlibrary{positioning, shapes, arrows.meta, calc}

\definecolor{darkred}{rgb}{0.6,0,0}
\definecolor{darkgreen}{rgb}{0,0.6,0}

\hypersetup{
    colorlinks=true,
    linkcolor=cyan,
    citecolor=magenta
}

\usepackage{orcidlink}

\begin{document}

\title{\textsc{BioMedVR}: \\ Confusion-Aware Mixture-of-Prompt Experts for \underline{Biomed}ical \underline{V}isual \underline{R}eprogramming}

\titlerunning{BioMedVR}

\author{Jiaxiang Liu\inst{1,2} \and
Tianxiang Hu\inst{2} \and
Juwei Guan\inst{3} \and
Yujie Wu\inst{4} \and
Yusong Wang\inst{5} \and
Yao Mu\inst{6} \and
Zuozhu Liu$^{*}$\inst{2} \and
Mingkun Xu$^{*}$\inst{1}}

\authorrunning{J.~Liu et al.}

\institute{Guangdong Institute of Intelligence Science and Technology, Zhuhai, China \and
Zhejiang University, Hangzhou, China \and
Southeast University, Nanjing, China \and
The Hong Kong Polytechnic University, Hong Kong SAR, China \and
Institute of Science Tokyo, Tokyo, Japan \and
Shanghai Jiao Tong University, Shanghai, China}

\maketitle

\renewcommand{\thefootnote}{\fnsymbol{footnote}}
\footnotetext[1]{Corresponding authors: Zuozhu Liu (zuozhuliu@intl.zju.edu.cn) and Mingkun Xu (xumingkun@gdiist.cn).}
\renewcommand{\thefootnote}{\arabic{footnote}}

\begin{abstract}
Recent advances in vision--language models (VLMs) such as CLIP have demonstrated strong generalization across natural-image domains.
However, adapting these models to biomedical imaging is non-trivial: full-model fine-tuning is computationally expensive, while medical data are often scarce and exhibit subtle, fine-grained inter-class differences, making parameter-efficient adaptation particularly critical.
Visual Reprogramming (VR) offers a parameter-efficient alternative by injecting learnable perturbations into the input space, but existing VR approaches for VLMs mainly focus on positive class prompts and overlook confusing negatives, leading to miscalibrated predictions in fine-grained medical scenarios.
We present BioMedVR, the first VR-based framework for biomedical imaging, enabling few-shot adaptation of pretrained VLMs through compact learnable VR modules.
To mitigate class confusion, we introduce a Confusion Minimization Mechanism that leverages LLM-generated confusion-aware attributes together with a Confusion-Suppression Loss to explicitly reduce false-positive alignment.
Moreover, the designed Mixture-of-Prompt Experts combines a positive expert for main-class discrimination and a negative expert for confusion suppression, balanced via adaptive gating.
Extensive experiments on 18 datasets---including 11 biomedical datasets and 7 natural image benchmarks---demonstrate that BioMedVR achieves superior accuracy and generalization, effectively bridging VR and VLMs in biomedical domains.
\end{abstract}

\section{Introduction}
\label{sec:intro}

Recent advances in vision-language models (VLMs), such as CLIP \cite{radford2021learning,xutowards,chenunderstanding,wangconnecting}, have opened new avenues for multimodal understanding. Unlike conventional supervised learning restricted to closed visual vocabularies, contrastive pretraining aligns image and text representations via natural language supervision, enabling open-set recognition. However, adapting large VLMs to specialized domains remains challenging: full model training is computationally prohibitive, and performance is highly sensitive to the design of both visual inputs and textual prompts \cite{chen2025r,wang2025fair}.

To alleviate this, prompt learning~\cite{zhou2022learning,zhou2022conditional,khattak2023maple,li2024promptkd} offers a parameter-efficient alternative by inserting learnable prompts into text tokens, image patches, attention layers, or cross-modal mappings \cite{chen2023understanding,oh2023blackvip,tsao2023autovp}.
Representative methods include CoOp~\cite{zhou2022learning}, CoCoOp \cite{zhou2022conditional}, and MaPLe \cite{khattak2023maple}, with BiomedCoOp~\cite{koleilat2025biomedcoop} extending this paradigm to biomedical imaging via LLM-guided prompt ensembles.

\begin{figure*}[t]
    \centering
    \includegraphics[width=\linewidth]{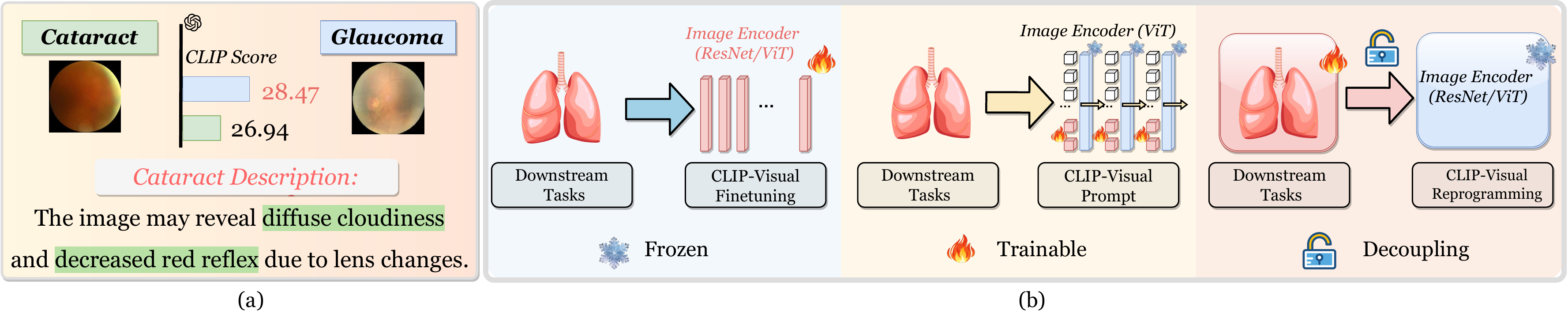}
    \caption{
(a) A cataract-specific description scores highly for glaucoma, exposing CLIP's semantic confusion and motivating the use of confusion-aware negative attributes to better separate similar diseases.
(b) Comparison of CLIP adaptation strategies.
(i) Finetuning updates all encoder parameters.
(ii) Visual Prompting injects learnable tokens within ViT layers.
(iii) VR learns lightweight input perturbations with frozen encoders, enabling efficient and architecture-agnostic adaptation.
While prompt learning is parameter-efficient, it typically requires access to model internals such as token embeddings or encoder interfaces. In contrast, VR operates in the input space, making it architecture-agnostic and better suited for privacy-sensitive medical deployments.
}
\label{fig:threeclip}

\end{figure*}

However, existing prompt-learning methods still depend on model internals and
struggle with fine-grained, modality-specific medical cues (\autoref{fig:threeclip}b).

An alternative and promising paradigm for adapting VLMs is {Visual Reprogramming (VR)} \cite{cai2024sample,chen2023understanding,chen2024model}.

Unlike prompt learning, which requires additional parameters and access to model internals, VR learns a trainable visual pattern in the input space, achieving architecture-agnostic transfer without modifying pretrained weights.
Originally explored for language reuse~\cite{hambardzumyan2021warp,vinod2020reprogramming}, the reprogramming paradigm has since been extended to graphs~\cite{jing2023deep}, vision~\cite{tsai2020transfer,chen2023understanding,caiattribute}, and speech modeling~\cite{yang2021voice2series,hung2023low}.
In the visual domain, VR overlays a learnable perturbation onto input images, which is optimized using task labels to repurpose frozen models.
When applied to VLMs~\cite{bahng2022exploring,chen2023understanding}, the perturbation functions as a \textit{input-space visual prompt}, jointly learned with textual templates for multimodal alignment.
Representative methods include VP~\cite{bahng2022exploring}, AR~\cite{tsai2020transfer,chen2023understanding}, and AttrVR~\cite{caiattribute}.
Among them, AttrVR~\cite{caiattribute} leverages descriptive and discriminative attributes to guide optimization, reducing intra-class variance and improving inter-class separability.

However, existing VR methods typically rely on a single shared visual pattern, limiting their capacity to capture heterogeneous semantics.
This limitation is especially pronounced in biomedical imaging, where class boundaries are subtle and visual cues are highly fine-grained (\autoref{fig:threeclip}a). As a result, directly transferring a single VR pattern from natural images often leads to overfitting to local textures or boundary artifacts, degrading diagnostic discrimination. These challenges motivate a confusion-aware, multi-expert VR framework tailored to biomedical tasks.

To address the above challenges, we propose \textsc{BioMedVR}, the first framework that introduces VR into biomedical image analysis. BioMedVR enables parameter-efficient adaptation of VLMs such as CLIP for few-shot medical image diagnosis (\autoref{tab:params}).

BioMedVR introduces a Confusion-aware Mixture-of-Prompt Experts \underline{(MoPE)} structure that explicitly models the relationship between main and confusing categories.

It comprises two complementary experts: a positive expert that enhances main-class recognition using positive attributes, and a negative expert that suppresses easily confused categories via large language model (LLM)-generated confusion-aware attributes, with their outputs adaptively fused through a learnable gating module.
Furthermore, we design a \underline{confusion suppression loss} that explicitly penalizes small similarity margins between true and confusing categories, effectively sharpening decision boundaries and improving training stability.

Extensive experiments on 11 biomedical datasets spanning 9 imaging modalities, together with 7 natural benchmarks, show that BioMedVR consistently surpasses existing prompt-learning and VR methods in both few-shot and zero-shot settings, delivering superior accuracy, robustness, and interpretability while using over \mbox{$500\times$} fewer trainable parameters than full CLIP fine-tuning.
These results validate the effectiveness of integrating VR into biomedical domains and establish BioMedVR as a lightweight yet powerful paradigm for explainable few-shot medical image diagnosis.
Our main contributions are as follows:
\begin{itemize}
    \item

    To our knowledge, we introduce VR into biomedical imaging for the first time, enabling a decoupled adaptation paradigm where downstream tasks are handled through VR rather than model modification, allowing parameter-efficient and privacy-preserving reuse of VLMs for few-shot medical image recognition.
    \item

    We propose a confusion-aware MoPE that decouples positive and negative experts: the positive expert exploits positive attributes for main-class recognition, while the negative expert uses confusion-aware attributes to suppress ambiguity. This confusion-aware design supports both zero-shot and few-shot medical adaptation.
    \item We design a confusion suppression loss that explicitly penalizes easily confused categories, reducing overconfidence and improving robustness. Extensive experiments on multiple medical datasets validate the superiority and interpretability of our approach.
\end{itemize}

\section{Related Work}

\subsection{VLM Adaptation and Prompt Learning.}

Large VLMs, such as CLIP~\cite{radford2021learning} and ALIGN~\cite{jia2021scaling}, have demonstrated strong zero-shot transfer by aligning visual and textual representations in a shared embedding space~\cite{liu2025kpl,liu2024vpl}.
To further adapt these pretrained models to downstream domains, recent works have explored various parameter-efficient strategies---such as prompt learning~\cite{zhou2022learning,khattak2023maple}, adapter tuning~\cite{houlsby2019parameter,liu2023parameter}, low-rank adaptation~\cite{hu2022lora,zanella2024low}, and VR~\cite{chen2023understanding,caiattribute}, which introduce learnable modules while keeping the backbone frozen.

Prompt learning has gained notable attention due to its simplicity and strong compatibility with VLMs.

Prompt-based tuning methods, including CoOp~\cite{zhou2022learning}, CoCoOp~\cite{zhou2022conditional}, MaPLe~\cite{khattak2023maple}, and PromptKD~\cite{li2024promptkd}, introduce learnable textual or visual tokens to adapt frozen backbones efficiently, while BiomedCoOp~\cite{koleilat2025biomedcoop} extends this paradigm to medical imaging via LLM-generated prompts. However, these approaches remain constrained by their reliance on {VPT} (visual prompt tuning) and linguistic assumptions, making them inadequate for modeling fine-grained, pixel-level medical features such as microvascular and tissue textures. In contrast, VR modifies the input space through lightweight, learnable perturbations, enabling architecture-agnostic, privacy-preserving, and interpretable adaptation across heterogeneous biomedical modalities.

\subsection{Visual Reprogramming.}

VR~\cite{elsayedadversarial,tsai2020transfer,chen2023understanding,bahng2022exploring,caiattribute} provides a parameter-efficient alternative for adapting pretrained models by learning input-space perturbations or task-specific visual prompts without modifying model parameters.
Unlike prompt-based or adapter-based methods that require access to model internals, VR operates at the input level and is therefore architecture-agnostic, compatible with both transformer and convolutional backbones \citep{yang2021voice2series,yang2023english}.
Recent variants such as AR~\cite{tsai2020transfer,chen2023understanding} and AttrVR~\cite{caiattribute} enhance interpretability by introducing attribute-guided or semantic-aware reprogramming, effectively bridging image--text alignment within VLMs.
Guided by LLMs, VR can optimize lightweight, input-level patterns to encode task-specific knowledge while keeping pretrained weights frozen, thereby reducing computational cost and avoiding catastrophic forgetting.

Given the heterogeneity of biomedical imaging modalities (e.g., ultrasound, MRI) and strict privacy constraints \cite{deng2025med}, VR offers an ideal mechanism for model reuse---it modifies only input patterns, ensuring data security and rapid adaptation across domains and organs \cite{liu2025kpl,koleilat2025biomedcoop}.
However, existing VR methods have mainly been validated on natural image benchmarks such as DTD, and Oxford-Pets, and typically rely on a single shared prompt for all categories.
This design limits flexibility and tends to amplify overconfidence in visually ambiguous biomedical scenarios characterized by high inter-class similarity.

\section{Methodology}
\label{sec:formatting}

\begin{figure*}[t]
    \centering
    \includegraphics[width=\linewidth]{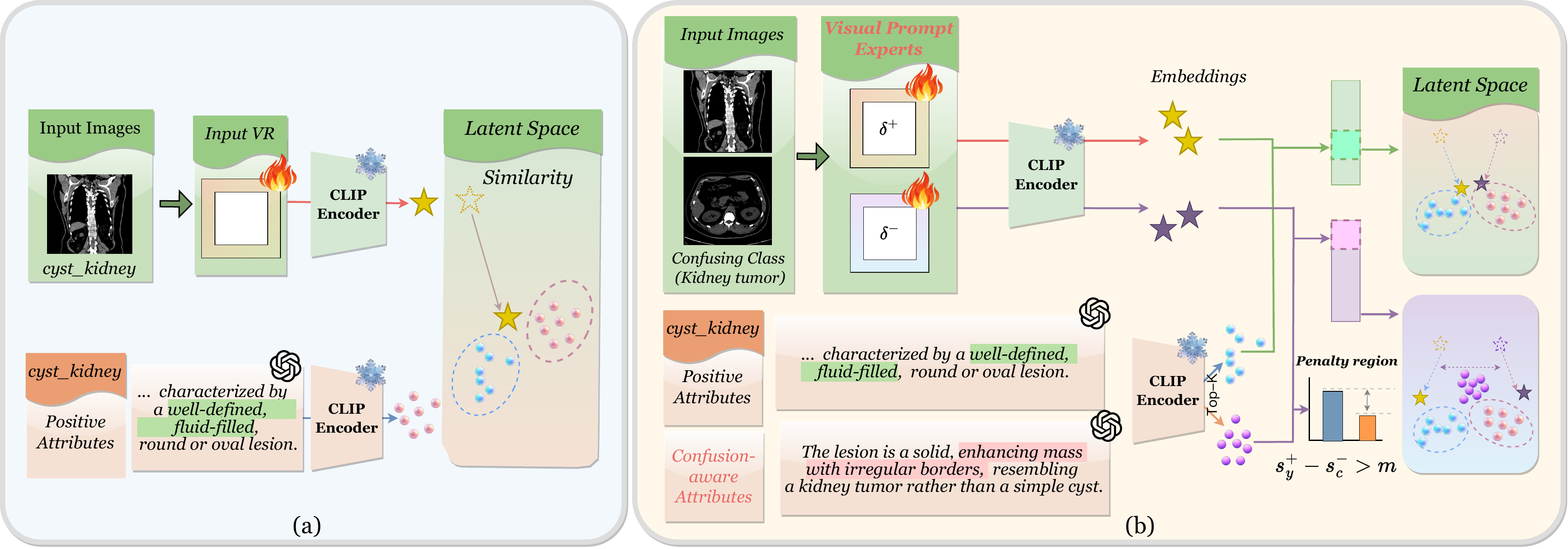}
    \caption{
{Comparison between Conventional VR and BioMedVR.}
(a) Conventional VR methods (e.g., AttrVR) apply a single visual prompt to align input images with positive textual attributes (e.g., ``well-defined, fluid-filled lesion''),
but fail to handle visually similar yet semantically incorrect classes such as \textit{kidney cyst} vs. \textit{kidney tumor}, leading to overlapping latent embeddings.
(b) In contrast, BioMedVR introduces a \textit{Confusion-aware MoPE} that employs visual experts---positive ($\delta^{+}$) and negative ($\delta^{-}$)---guided by descriptive and confusion-aware attributes respectively.
The penalty region suppresses inter-class confusion in the embedding space.
This design explicitly models fine-grained semantic conflicts, yielding more discriminative and well-separated representations in medical image diagnosis.}
    \label{fig:pipeline}

\end{figure*}

\subsection{Preliminaries}
Given a pretrained VLM, such as CLIP~\cite{radford2021learning}, we denote its frozen visual and textual encoders as $f_v(\cdot)$ and $f_t(\cdot)$, respectively.
For an input image $x \in \mathbb{R}^{H\times W\times 3}$ and a category label $y \in \mathcal{Y}$, the corresponding textual prompt or attribute description $t_y$ is first embedded as
\begin{equation}
    z_v = \frac{f_v(x)}{\|f_v(x)\|}, \quad
    z_t = \frac{f_t(t_y)}{\|f_t(t_y)\|},
\end{equation}
where $z_v, z_t \in \mathbb{R}^d$ are the normalized feature embeddings in the shared multimodal space.
The similarity score between an image $x$ and a text $t_y$ is computed as
\begin{equation}
    s(x, t_y) = \exp(\tau \cdot z_v^\top z_t),
\end{equation}
where $\tau = \exp(\text{logit\_scale})$ is a learnable temperature parameter from CLIP.

\vspace{4pt}
\noindent\textbf{Visual Reprogramming.}
In VR~\cite{caiattribute}, instead of tuning model parameters, a learnable visual prompt $\delta$ is introduced at the input level to adapt the pretrained model to a new target domain.
The reprogrammed image is formulated as
\begin{equation}
    \tilde{x} = \mathcal{R}(x; \delta) = x \odot M + \delta \odot (1 - M),
\end{equation}
where $M$ denotes a binary mask controlling the region to be perturbed.
The objective of VR is to learn $\delta$ such that the reprogrammed input $\tilde{x}$ can be correctly aligned with textual representations of the target label $y$:
\begin{equation}
    \mathcal{L}_{vr} = -\frac{1}{N} \sum_{i=1}^N \log
    \frac{\exp(\tau \cdot f_v(\tilde{x}_i)^\top f_t(t_{y_i}))}
    {\sum_{y' \in \mathcal{Y}} \exp(\tau \cdot f_v(\tilde{x}_i)^\top f_t(t_{y'}))}.
\end{equation}

\vspace{4pt}
\noindent\textbf{Attribute-Guided Reprogramming.}
In practice, each class $y$ can be described by a set of textual attributes
$\mathcal{T}_y = \{\mathbf{t}_{y,1}, \mathbf{t}_{y,2}, \dots, \mathbf{t}_{y,K}\}$
representing \textit{descriptive} and \textit{discriminative} semantics \cite{caiattribute}.
Accordingly, the VR objective can be reformulated as an attribute-level similarity aggregation:
\begin{equation}
\small
    \mathcal{L}_{attr} = -\frac{1}{N} \sum_{i=1}^N \log
    \frac{\sum_{t \in \mathcal{T}_{y_i}} \exp(\tau \cdot f_v(\tilde{x}_i)^\top f_t(t))}
    {\sum_{y'\in \mathcal{Y}} \sum_{t' \in \mathcal{T}_{y'}} \exp(\tau \cdot f_v(\tilde{x}_i)^\top f_t(t'))}.
\end{equation}
This formulation bridges image-text alignment at the attribute level, serving as the foundation for the proposed confusion-aware MoPE introduced in Section~\ref{sec:MoPE}.

\subsection{Analysis of the VR in Biomedical Imaging}
\label{sec:analysis}

In the VR paradigm, the optimization aims to learn an input-level perturbation $\delta$ that maximizes the similarity between the reprogrammed image $\tilde{x}=x+\delta$ and its corresponding text $t_y$ while minimizing similarity to other classes:
\begin{equation}
\min_{\delta} \mathcal{L}_{vr} = -\log
\frac{\exp(\tau \cdot f_v(\tilde{x})^\top f_t(t_y))}
{\sum_{y' \in \mathcal{Y}} \exp(\tau \cdot f_v(\tilde{x})^\top f_t(t_{y'}))}.
\label{eq:vr_loss}
\end{equation}

\begin{wrapfigure}{l}{0.52\linewidth}
\vspace{-6pt}
\centering
\includegraphics[width=0.95\linewidth]{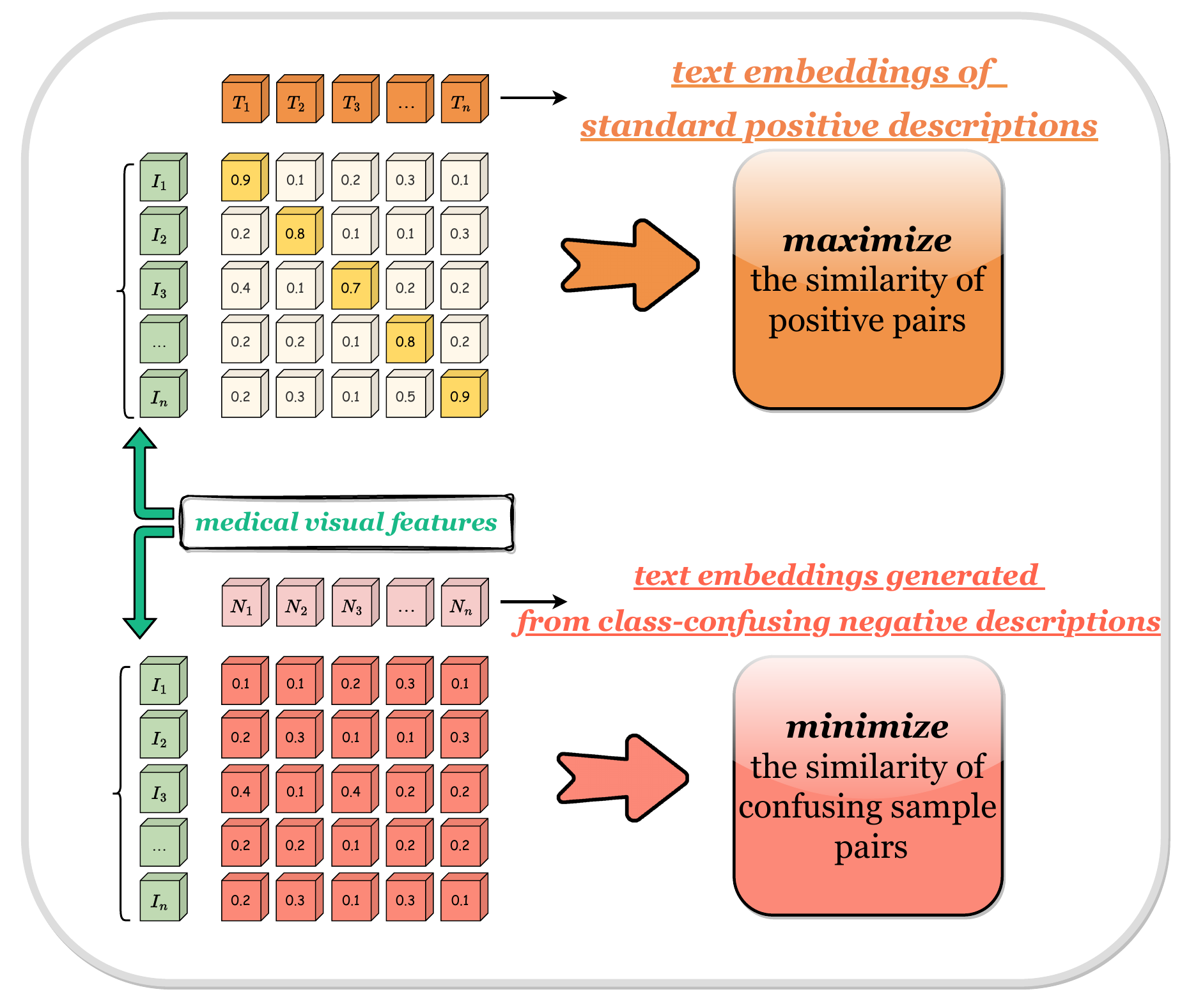}
\vspace{-8pt}
\caption{\footnotesize
\textbf{Confusion-aware optimization objective.}
BioMedVR maximizes alignment with positive attributes while suppressing confusing negative attributes generated from visually similar but incorrect categories, enlarging the semantic margin and reducing inter-class confusion.}
\label{fig:confuse-obj}
\vspace{-10pt}
\end{wrapfigure}

However, directly applying this approach to biomedical image tasks raises two notable challenges. First, in biomedical imaging, categories often exhibit high inter-class similarity and low intra-class variance. Formally, for two classes $y_i, y_j \in \mathcal{Y}$ with embeddings $f_t(t_{y_i})$ and $f_t(t_{y_j})$, their cosine similarity $f_t(t_{y_i})^\top f_t(t_{y_j})  \!\approx\! 1$, especially when $y_i$ and $y_j$ correspond to subtle disease subtypes (e.g., \textit{glaucoma} vs. \textit{cataract}).
As a result, the softmax distribution in Eq.~\eqref{eq:vr_loss} becomes flat and the gradients of the loss diminish, providing little discriminative signal for optimizing $\delta$.

Second, the single shared perturbation $\delta$ is globally applied across all classes,
implicitly assuming a one-to-one alignment between visual prompts and class semantics.
In practice, this can be decomposed as
\begin{equation}
f_v(\tilde{x}) = f_v(x + \delta) \approx f_v(x) + J_{f_v}(x) \cdot \delta,
\end{equation}
where $J_{f_v}(x)$ is the Jacobian of the visual encoder.
For heterogeneous biomedical modalities (e.g., MRI, CT), $J_{f_v}(x)$ can vary drastically across domains, causing a single $\delta$ to misalign with semantic boundaries and amplify confusion between visually similar categories.

These observations motivate the development of \textsc{BioMedVR}, summarized in \textcolor{red}{Algorithm}~\ref{alg:BioMedVR}, to overcome the challenges of biomedical visual reprogramming.

\subsection{Confusion-aware MoPE}
\label{sec:MoPE}

We extend the visual reprogramming paradigm by introducing a \textit{Confusion-aware MoPE}, which explicitly disentangles discriminative and confusable semantics via dual experts and margin-based optimization, as shown in \autoref{fig:pipeline}.

Given an image $x$ with label $y$, two prompt experts---\textit{positive} ($\delta^{+}$) and \textit{negative} ($\delta^{-}$)---generate reprogrammed inputs:
\begin{equation}
\tilde{x}^{+}=x+\delta^{+}, \qquad \tilde{x}^{-}=x+\delta^{-},
\end{equation}
whose normalized embeddings are
\begin{equation}
z^{+}=\frac{f_v(\tilde{x}^{+})}{\|f_v(\tilde{x}^{+})\|}, \quad
z^{-}=\frac{f_v(\tilde{x}^{-})}{\|f_v(\tilde{x}^{-})\|}.
\end{equation}

The positive expert leverages descriptive and discriminative textual embeddings $\mathcal{T}^{+}_y$ \cite{caiattribute},
while the negative expert learns from confusion-aware textual attributes $\mathcal{T}^{-}_y$ automatically generated by an LLM.
For each class $y$, we query the LLM with \textit{``Generate the most visually confusing negative descriptions for class [Class Name]''}, obtaining
\begin{equation}
\mathcal{T}^{-}_y=\{\,t^{-}_i\,\}_{i=1}^{N_c}, \quad t^{-}_i\sim\mathrm{LLM}(y),
\end{equation}
where $N_c$ denotes the number of confusion-aware attributes.

During optimization, the experts respectively compute their alignment scores:
\begin{equation}
\small
s^{+}_y=\frac{\tau}{k}\!\sum_{t\in\mathrm{Top}_k(\mathcal{T}^{+}_y)}\!z^{+\top}f_t(t), \quad
s^{-}_y=\frac{\tau}{k}\!\sum_{t\in\mathrm{Top}_k(\mathcal{T}^{-}_y)}\!z^{-\top}f_t(t),
\end{equation}
where $\tau$ denotes the temperature and $k$ controls top-$k$ attribute selection.
The fused logit combines both experts through an adaptive gating mechanism:
\begin{equation}
s_y = g^{+} s^{+}_y + g^{-} s^{-}_y, \qquad
[g^{+}, g^{-}] = \mathrm{softmax}(w),
\end{equation}
where $w \in \mathbb{R}^2$ is a learnable gating vector shared across all samples.
Each element of $w$ corresponds to the confidence weight of the positive or negative expert,
and is automatically optimized during training to balance discriminative enhancement and confusion suppression.

Instead of relying on any single logit, our MoPE formulation naturally induces a two-stream separation between a discriminative pathway $s^{+}_y$ and a confusion pathway $\max_{c\neq y}s^{-}_c$. This separation allows downstream objectives (Sec. \ref{sec:confusion}) to directly manage the semantic margin:

\begin{equation}
\Gamma(x,y)=s^{+}_y-\max_{c\neq y}s^{-}_c \;\;\uparrow,
\end{equation}

which quantifies how well the positive expert distinguishes the true class from the most confusable negatives. Our confusion-suppression loss explicitly enlarges this margin during training.
{Intuitively, $\max_{c\neq y}s^{-}_c$ can be viewed as the score of the most plausible alternative diagnosis candidate, so enlarging $\Gamma(x,y)$ directly promotes differential-diagnosis-like separation.}

\subsection{Confusion-Suppression Objective}

\label{sec:confusion}

The Confusion-Suppression (CS) loss is defined as:
\begin{equation}
\mathcal{L}_{cs} =
\mathbb{E}_{(x,y)}\!
\left[\, \max\!\big(0,\, \max_{c\neq y}s^{-}_{c} - s^{+}_{y} + m \big)\, \right],
\end{equation}

where $m$ is a margin hyperparameter that defines the penalty region for confusion. Minimizing $\mathcal{L}$ enforces the constraint:
\begin{equation}
s^{+}_y - \max_{c\neq y}s^{-}_c \ge m,
\end{equation}
which explicitly enlarges the semantic margin between discriminative and confusable categories. This loss activates only when confusion occurs, preventing unnecessary suppression of legitimately ambiguous samples (\autoref{fig:confuse-obj}).
{By focusing on the strongest confusable alternative, the objective targets the most harmful errors in fine-grained biomedical recognition.}

The final training objective combines standard cross-entropy with confusion suppression:
\begin{equation}
\mathcal{L} = \mathcal{L}_{CE} + \beta\,\mathcal{L}_{cs},
\end{equation}
where $\beta$ balances classification accuracy and confusion robustness. This constraint reshapes the decision boundary by suppressing high-similarity negatives, yielding more stable logits and calibrated predictions in biomedical domains.

\begin{algorithm}[t]
\caption{\textbf{Training Procedure of BioMedVR}}
\label{alg:BioMedVR}
\begin{algorithmic}[1]
\Require
Frozen VLM encoders $(f_v, f_t)$; dataset $\mathcal{D}=\{(x_i, y_i)\}$; \\
Descriptive \& discriminative (positive) text sets $\mathcal{T}^{+}_y$; Confusion-aware text sets $\mathcal{T}^{-}_y$; \\
Hyperparameters $\beta, m, k, lr, Epoch$.
\Ensure
Optimized visual reprogramming parameters $\{\delta^{+}, \delta^{-}, w\}$.
\State Initialize positive/negative experts $\{\delta^{+}, \delta^{-}\}$ and gating logits $w=[1,1]$.
\For{epoch $=1$ to $Epoch$}
    \For{each mini-batch $(x, y)$ in $\mathcal{D}$}
        \State \textbf{Reprogramming:}  \small $\tilde{x}^{+}=x+\delta^{+}, \quad \tilde{x}^{-}=x+\delta^{-}$.
        \State \textbf{Embedding:} $z^{+}=\frac{f_v(\tilde{x}^{+})}{\|f_v(\tilde{x}^{+})\|}, \;
        z^{-}=\frac{f_v(\tilde{x}^{-})}{\|f_v(\tilde{x}^{-})\|}$.
        \State \textbf{Gating:} $[g^{+}, g^{-}] = \mathrm{softmax}(w)$.
        \State \textbf{Expert Logits:}
\State {\small $
s^{+}_y=\frac{\tau}{k}\!\!\sum_{t\in\mathrm{Top}_k(\mathcal{T}^{+}_y)}\!z^{+\top}f_t(t),$  \\
\ \ \quad \quad \quad $s^{-}_y=\frac{\tau}{k}\!\!\sum_{t\in\mathrm{Top}_k(\mathcal{T}^{-}_y)}\!z^{-\top}f_t(t)
$}
        \State \textbf{Fusion:} \small $s_y = g^{+} s^{+}_y + g^{-} s^{-}_y,\quad
        p(y|x)=\frac{e^{s_y}}{\sum_c e^{s_c}}.$
        \State \textbf{Loss:}
        \footnotesize $
        \mathcal{L} = -\log p(y|x)
        + \beta [\max_{c\neq y}s^{-}_c - s^{+}_y + m]_+.
        $
        \State Update $\{\delta^{+}, \delta^{-}, w\}$ via SGD with cosine annealing.
    \EndFor
\EndFor
\State \Return $\{\delta^{+}, \delta^{-}, w\}$.
\end{algorithmic}

\end{algorithm}

\section{Experiments}

\begin{table*}[t]
\centering
\setlength{\extrarowheight}{0pt}
\addtolength{\extrarowheight}{\aboverulesep}
\addtolength{\extrarowheight}{\belowrulesep}
\setlength{\aboverulesep}{0pt}
\setlength{\belowrulesep}{0pt}
\caption{
Accuracy (\%) comparison of zero-shot and few-shot methods on 16-shot biomedical
classification tasks using ViT-B/16--based CLIP as the pretrained model
(the highest values are in \textbf{bold}; ZS denotes the zero-shot setting).
In the few-shot block, rows with the light-peach background indicate VR-based methods,
while the remaining rows correspond to prompt-learning approaches.
}
\resizebox{\textwidth}{!}{
\begin{tabular}{lcccccccccccc}
\toprule
\textbf{Method} & BUSI & Knee X-ray & Kvasir & LungColon & OCTMNIST & BTMRI & CHMNIST & COVID-19 & CT-Kidney & DermaMNIST & Retina & {Avg.} \\
\midrule
\multicolumn{13}{c}{{\cellcolor[rgb]{0.925,0.925,0.925}}\textit{Zero-shot}} \\
CLIP \cite{radford2021learning} \textit{[ICML 2021]} & 30.5\% & 34.7\% & 18.7\% & 20.4\% & 33.2\% & 27.3\% & 32.0\% & 36.4\% & 38.1\% & 13.0\% & 25.5\% & 28.1\% \\
AttrVR (ZS) \textit{[ICLR 2025]} \cite{caiattribute} & 26.3\% & 38.6\% & 13.8\% & 40.8\% & 26.6\% & 44.5\% & 20.2\% & 07.6\% & 30.1\% & 09.7\% & 43.1\% & 27.4\% \\
BioMedVR (ZS) & {46.6\%} & {38.3\%} & {14.1\%} & {39.3\%} & {24.4\%} & {46.2\%} & {21.3\%} & {28.4\%} & {24.5\%} & {13.9\%} & {49.6\%} & {31.6\%} \\
BioMedCLIP (ZS) \textit{[NEJM AI 2025]} \cite{zhang2025multimodal} & 54.2\% & 25.5\% & 51.5\% & 38.1\% & 47.0\% & 57.2\% & 23.2\% & 59.4\% & 41.9\% & 19.3\% & 32.1\% & 40.9\% \\
BioMedVR (ZS, BioMedCLIP) & 54.2\% & 36.4\% & 62.9\% & 67.0\% & 64.7\% & 58.1\% & 36.2\% & 61.5\% & 32.9\% & 04.8\% & 45.0\% & 47.6\% \\
\midrule
\multicolumn{13}{c}{{\cellcolor[rgb]{0.925,0.925,0.925}}\textit{Few-shot}} \\
CoOp \citep{zhou2022learning} \textit{[IJCV'22]}  & 62.3\% & 27.1\% & 74.8\% & 89.6\% & 67.9\% & 79.2\% & {76.7\%} & 73.6\% & \textbf{81.8}\% & 43.0\% & 27.1\% & 63.9\% \\

CoCoOp \citep{zhou2022conditional} \textit{[CVPR'22]}  & 64.4\% & 30.0\% & 77.5\% & 89.4\% & 68.5\% & 80.4\% & {71.5\%} & 76.2\% & 78.0\% & 44.0\% & 52.5\% & 66.6\% \\

BioMedCoOp \citep{koleilat2025biomedcoop} \textit{[CVPR'25]}  & 69.4\% & 36.8\% & 77.7\% & 91.5\% & 72.5\% & 81.0\% & {76.9\%} & 77.0\% & {80.9}\% & 59.9\% & 59.9\% & 71.2\% \\
\rowcolor[rgb]{0.994,0.966,0.949}
VP \citep{bahng2022exploring} & 70.8\% & 41.6\% & 75.2\% & 90.8\% & 65.1\% & 65.9\% & 70.8\% & 67.8\% & 67.8\% & 64.9\% & 70.4\% & 68.3\% \\
\rowcolor[rgb]{0.994,0.966,0.949}
AR \cite{chen2023understanding} \textit{[CVPR'23]} & 75.4\% & 39.4\% & 79.3\% & 94.3\% & 77.2\% & 75.7\% & 83.9\% & 70.1\% & 72.5\% & 58.0\% & 73.3\% & 72.6\% \\

\rowcolor[rgb]{0.994,0.966,0.949}
AttrVR \citep{caiattribute} \textit{[ICLR'25]}
& 78.0\% & 33.5\% & 79.4\% & 93.8\% & \textbf{80.4}\% & 76.5\% & \textbf{85.3}\% & 71.0\% & 71.6\% & 61.6\% & 71.5\% & 73.0\% \\

\rowcolor[rgb]{0.949,0.973,0.988}
{BioMedVR (Ours)} & \textbf{82.6\%} & \textbf{45.7\%} & \textbf{80.2\%} & \textbf{94.7\%} & {80.3\%} & \textbf{81.7\%} & 84.5\% & \textbf{77.4\%} & {74.0\%} & \textbf{65.3\%} & \textbf{74.1\%} & {\textbf{76.4\%}} \\
\bottomrule
\end{tabular}}
\label{exp:main}

\end{table*}

\subsection{Experimental Setup}

\noindent\textbf{Datasets.}
We comprehensively evaluate BioMedVR across 11 biomedical imaging datasets encompassing 10 organs and 9 imaging modalities, including Computerized Tomography (CTKidney~\cite{ctkidney}), Dermatoscopy (DermaMNIST~\cite{dermamnist1,dermamnist2}), Endoscopy (Kvasir~\cite{kvasir}), Fundus Photography (RETINA~\cite{retina1,retina2}), Histopathology (LungColon~\cite{LC25000}, CHMNIST~\cite{chmnist}), Magnetic Resonance Imaging (BTMRI~\cite{btmri}), Optical Coherence Tomography (OCTMNIST~\cite{octmnist}), Ultrasound (BUSI~\cite{busi}), and X-Ray (COVID-19~\cite{covid}, Knee X-ray~\cite{kneexray}).
This collection covers a broad range of diagnostic challenges---from high-contrast radiology to fine-grained histopathology---providing a rigorous testbed for cross-modal generalization.
In addition, we further validate BioMedVR on seven natural image benchmarks---Caltech101~\cite{caltech101}, Food101~\cite{food101}, DTD~\cite{dtd}, OxfordPets~\cite{parkhi2012cats}, Flowers102~\cite{flowers}, EuroSAT~\cite{eurosat}, and UCF101~\cite{ucf101}---to assess its generalization beyond the biomedical domain.
Detailed dataset splits and task configurations are provided in the supplementary material.

\noindent\textbf{Implementation Details.}
Encoders ($f_v$, $f_t$) are kept frozen, while only the $\{\delta^{+}, \delta^{-}\}$ and the gating vector $w$ are trainable.
We follow a few-shot learning setting with 4, 8, and 16 labeled samples per class, consistent with prior works~\cite{caiattribute}.

Unless otherwise noted, we adopt the 16-shot setting with the remaining samples for testing and use GPT-4.1 as the default LLM.
We also evaluate GPT-4o-mini, GPT-5-mini, and GPT-5 to study the impact of different LLMs on attribute generation.
For generating positive attributes, we follow the procedure described in \cite{caiattribute}. $N_c$ and the choice of top-$k$ follow the settings in \cite{caiattribute}.

Zero-shot BioMedVR combines positive cues with confusion-aware attribute suppression, using fused top-k similarities to select the most compatible class without training.
Training is performed with CLIP backbones (ViT-B/16, ViT-B/32, and RN50) using SGD (initial learning rate 40, momentum 0.9) and cosine annealing scheduling for 400 epochs with a batch size of 512.
The confusion margin $m$ and loss weight $\beta$ are set to 0.5 and 0.3, respectively, based on hyperparameter search.
{We report training/inference efficiency, parameter overhead, and memory comparisons to PEFT baselines in the supplementary material.}
All experiments are conducted on NVIDIA H20 GPUs under identical data splits for fair comparison.

\begin{figure}[t]
    \centering
    \includegraphics[width=\linewidth]{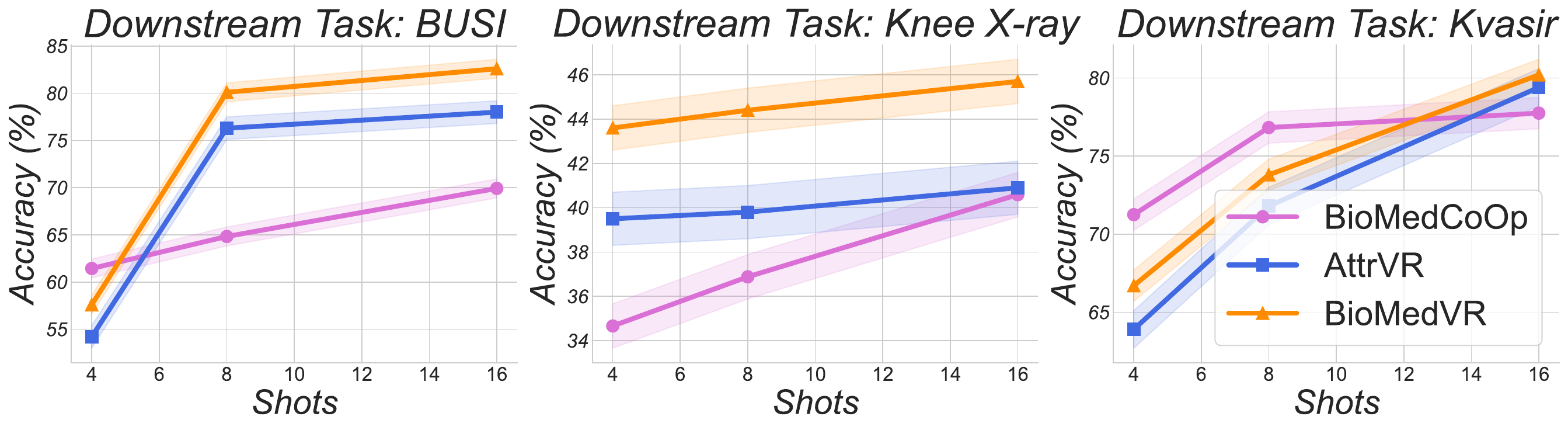}
    \caption{
    {Sample efficiency comparison across few-shot settings.}
Performance comparison of BioMedVR with AttrVR \citep{caiattribute} and BioMedCoOp \citep{koleilat2025biomedcoop} on three representative biomedical datasets.
    }
    \label{fig:sample_efficiency}

\end{figure}

\subsection{Comparison with State-of-the-Art (SOTA)}

We evaluate BioMedVR under both \textit{few-shot} and \textit{zero-shot} settings for a comprehensive comparison.
In the few-shot setting, we compare BioMedVR with SOTA VR-based VLM adaptation methods, including VP~\cite{bahng2022exploring}, which overlays learnable reprogramming patterns on resized images; AR~\cite{tsai2020transfer,chen2023understanding}, which pads structured visual patterns around image borders; and

AttrVR~\cite{caiattribute} performs attribute-based reprogramming using descriptive and discriminative textual cues to capture shared and class-specific semantics (GPT-5 is used since GPT-3.5 API is unavailable).

To assess the adaptability of our method, we further compare with prompt-based baselines built on the BiomedCLIP backbone~\cite{zhang2025multimodal}, including CoOp~\cite{zhou2022learning}, CoCoOp~\cite{zhou2022conditional}, and BiomedCoOp~\cite{koleilat2025biomedcoop}.
For zero-shot evaluation, we include the original CLIP, the zero-shot variant of AttrVR~\cite{caiattribute}, and zero-shot BioMedVR, which combines positive and confusion-aware textual cues without any fine-tuning.

\begin{table*}[t]
\centering
\begin{minipage}[t]{0.48\textwidth}
  \centering
  \captionof{table}{Training cost of different VR methods, using the ViT-B16-based CLIP on BUSI.}
  \resizebox{\linewidth}{!}{
  \begin{tabular}{ccccc}
  \toprule
   & AR & AttrVR & {\cellcolor[rgb]{0.953,0.973,0.988}}BioMedVR & CLIP \\
  \midrule
  Parameter Number & 0.15M & 0.15M & {\cellcolor[rgb]{0.953,0.973,0.988}}0.30M & \textasciitilde{}150M \\
  Training Time / Epoch (s) & 7.12 & 6.55 & {\cellcolor[rgb]{0.953,0.973,0.988}}9.10 & -- \\
  Performance (\%) & 72.6 & 73.0 & {\cellcolor[rgb]{0.953,0.973,0.988}}\textbf{76.4} & 27.3 \\
  \bottomrule
  \end{tabular}}
  \label{tab:params}
\end{minipage}\hfill
\begin{minipage}[t]{0.48\textwidth}
  \centering
  \captionof{table}{Few-shot and zero-shot performance on seven natural-image benchmarks (\%).}
  \resizebox{\linewidth}{!}{
  \begin{tabular}{ccccc|cc}
  \toprule
  \textbf{Dataset} & AR & VP & AttrVR & BioMedVR & CLIP & BioMedVR (ZS) \\
  \midrule
   & \multicolumn{4}{c|}{\textit{few-shot}} & \multicolumn{2}{c}{\textit{zero-shot}} \\
  \midrule
  UCF101 & 77.7 & 73.2 & 78.6 & {\cellcolor[rgb]{0.953,0.973,0.988}}\textbf{80.1} & 59.4 & {\cellcolor[rgb]{0.994,0.966,0.949}}\textbf{62.7} \\
  Caltech101 & 95.5 & 93.8 & \textbf{95.9} & {\cellcolor[rgb]{0.953,0.973,0.988}}95.7 & 85.4 & {\cellcolor[rgb]{0.994,0.966,0.949}}\textbf{92.1} \\
  Food101 & 85.4 & 82.3 & 86.0 & {\cellcolor[rgb]{0.953,0.973,0.988}}\textbf{86.0} & 72.8 & {\cellcolor[rgb]{0.994,0.966,0.949}}\textbf{78.9} \\
  DTD & 59.8 & 61.0 & 65.9 & {\cellcolor[rgb]{0.953,0.973,0.988}}\textbf{66.0} & 37.4 & {\cellcolor[rgb]{0.994,0.966,0.949}}\textbf{47.3} \\
  EuroSAT & 91.6 & 89.5 & 93.4 & {\cellcolor[rgb]{0.953,0.973,0.988}}\textbf{93.9} & 21.2 & {\cellcolor[rgb]{0.994,0.966,0.949}}21.0 \\
  Oxford-Pets & 92.6 & 91.3 & 93.1 & {\cellcolor[rgb]{0.953,0.973,0.988}}\textbf{93.9} & 77.6 & {\cellcolor[rgb]{0.994,0.966,0.949}}\textbf{88.4} \\
  Oxford-Flowers & 86.2 & 83.3 & \textbf{93.1} & {\cellcolor[rgb]{0.953,0.973,0.988}}93.0 & 59.6 & {\cellcolor[rgb]{0.994,0.966,0.949}}\textbf{72.2} \\
  \bottomrule
  \end{tabular}}
  \label{exp:nature}
\end{minipage}
\end{table*}

\noindent\textbf{Few-shot Setting:}
We compare BioMedVR with existing reprogramming methods using CLIP with the ViT-B/16.
As shown in \autoref{exp:main}, BioMedVR consistently outperforms baseline VR approaches, including AttrVR, VP, and AR, achieving an average improvement of over 3\% across 11 biomedical imaging datasets.
The performance gain is particularly notable on confusion-prone datasets such as \textit{Knee X-ray} and \textit{DermaMNIST}, demonstrating that explicit confusion modeling effectively enhances robustness and generalization.

In comparison with prompt-learning-based methods, our BioMedVR also surpasses CoOp, CoCoOp, and BiomedCoOp on most datasets.

Although CoOp and BiomedCoOp perform slightly better on \textit{CTKidney}, this likely stems from their use of the domain-specialized BiomedCLIP ViT-B/16.
Nevertheless, despite leveraging a general-domain CLIP backbone, BioMedVR still achieves superior average accuracy, underscoring its strong adaptability and effectiveness in few-shot medical scenarios.

We further evaluate BioMedVR on seven natural image datasets (\autoref{exp:nature}).
It achieves the best or comparable results across all benchmarks, notably improving over AR, VP, and AttrVR on challenging datasets such as \textit{UCF101}.
These results confirm that the proposed confusion-aware reprogramming generalizes effectively beyond medical imaging, demonstrating strong cross-domain adaptability.

\noindent\textbf{Zero-shot Setting.}
We further evaluate BioMedVR under the zero-shot setting using both CLIP and BiomedCLIP backbones.

As shown in \autoref{exp:main}, BioMedVR (ZS) improves the average zero-shot accuracy over standard CLIP from 28.1\% to 31.6\%, and reaches 47.6\% when built on BiomedCLIP, demonstrating strong overall performance across the 11 biomedical datasets.
Notably, significant gains are observed on complex modalities such as Knee X-ray (+10.9\%) and LungColon (+28.9\%), indicating that the confusion-aware attributes effectively enhance cross-modal generalization even without fine-tuning.
Moreover, when extended to natural image benchmarks (\autoref{exp:nature}), BioMedVR (ZS) achieves clear improvements over CLIP in zero-shot classification, particularly on \textit{Caltech101} (+6.7\%) and \textit{Oxford-Pets} (+10.8\%).
These results demonstrate that BioMedVR leverages LLM-guided descriptive and confusion-aware attributes to perform reliable zero-shot reasoning, narrowing the performance gap between zero-shot and few-shot paradigms while maintaining strong transferability.

\begin{figure*}[t]
\centering
\begin{minipage}[t]{0.48\textwidth}
  \centering
  \includegraphics[height=3.2cm]{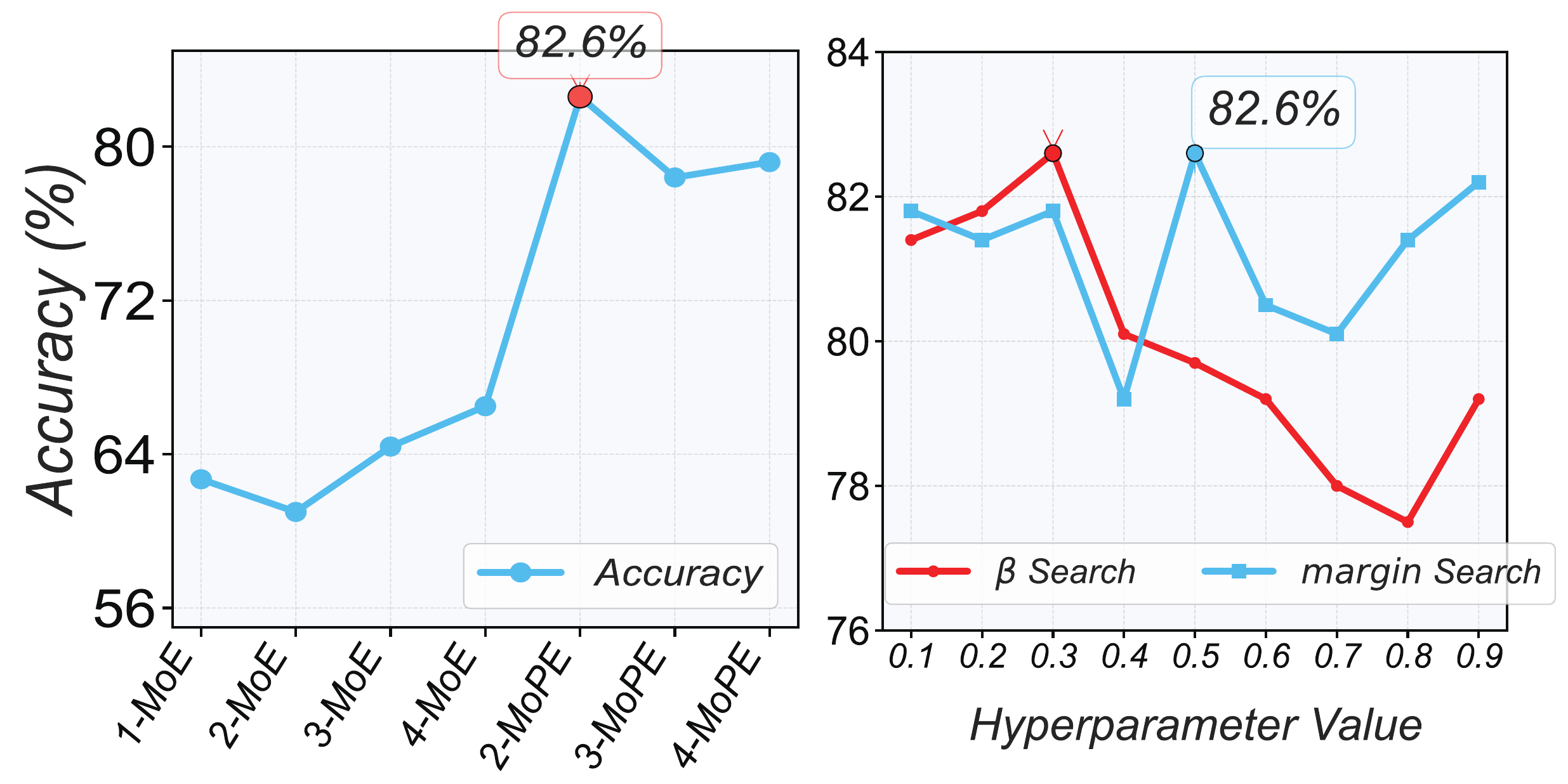}
  \caption{\textbf{Ablation on MoPE and hyperparameters.}
  2-MoPE yields the highest accuracy (82.6\%), with the optimum at $\beta{=}0.3$, $m{=}0.5$.}
  \label{fig:params}
\end{minipage}\hfill
\begin{minipage}[t]{0.48\textwidth}
  \centering
  \includegraphics[height=3cm]{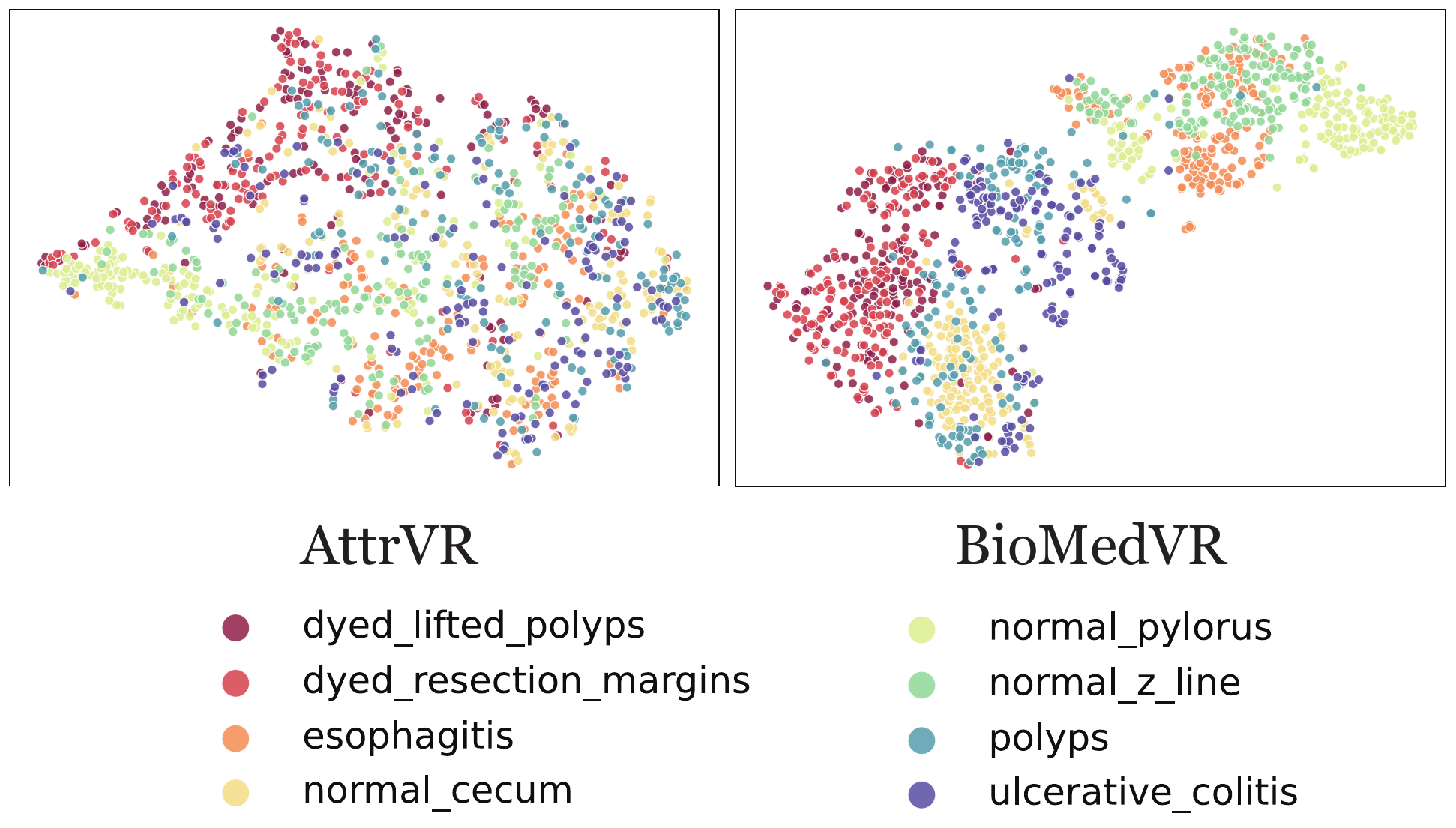}
  \caption{\textbf{t-SNE visualization on Kvasir.}
  BioMedVR forms more compact and better-separated clusters than AttrVR.}
  \label{fig:tsne_kvasir}
\end{minipage}
\end{figure*}

\subsection{Analysis and Ablation Study}
\textbf{Results of Sample Efficiency.}

We evaluate BioMedVR under 4-, 8-, and 16-shot settings.
As shown in \autoref{fig:sample_efficiency}, BioMedVR outperforms CoOp and AttrVR across most configurations and remains competitive even with only 4 samples per class.
Its performance saturates faster with increasing samples, demonstrating superior sample efficiency and effective utilization of limited supervision in data-scarce biomedical scenarios.

\noindent\textbf{Results on Different Backbones.}

We evaluate the generality of BioMedVR across multiple CLIP backbones, including ViT-B/16, ViT-B/32, and RN50.
As shown in \autoref{exp:main} and \autoref{exp:abl}, BioMedVR consistently outperforms its corresponding baselines on all backbones, demonstrating strong adaptability to both transformer- and convolution-based architectures.
Compared with the baseline AttrVR, BioMedVR achieves an average improvement of +2.6\% on RN50 and +1.5\% on ViT-B/32, confirming its robustness even under weaker feature representations.

\begin{figure*}[!htbp]
    \centering
    \includegraphics[width=\linewidth]{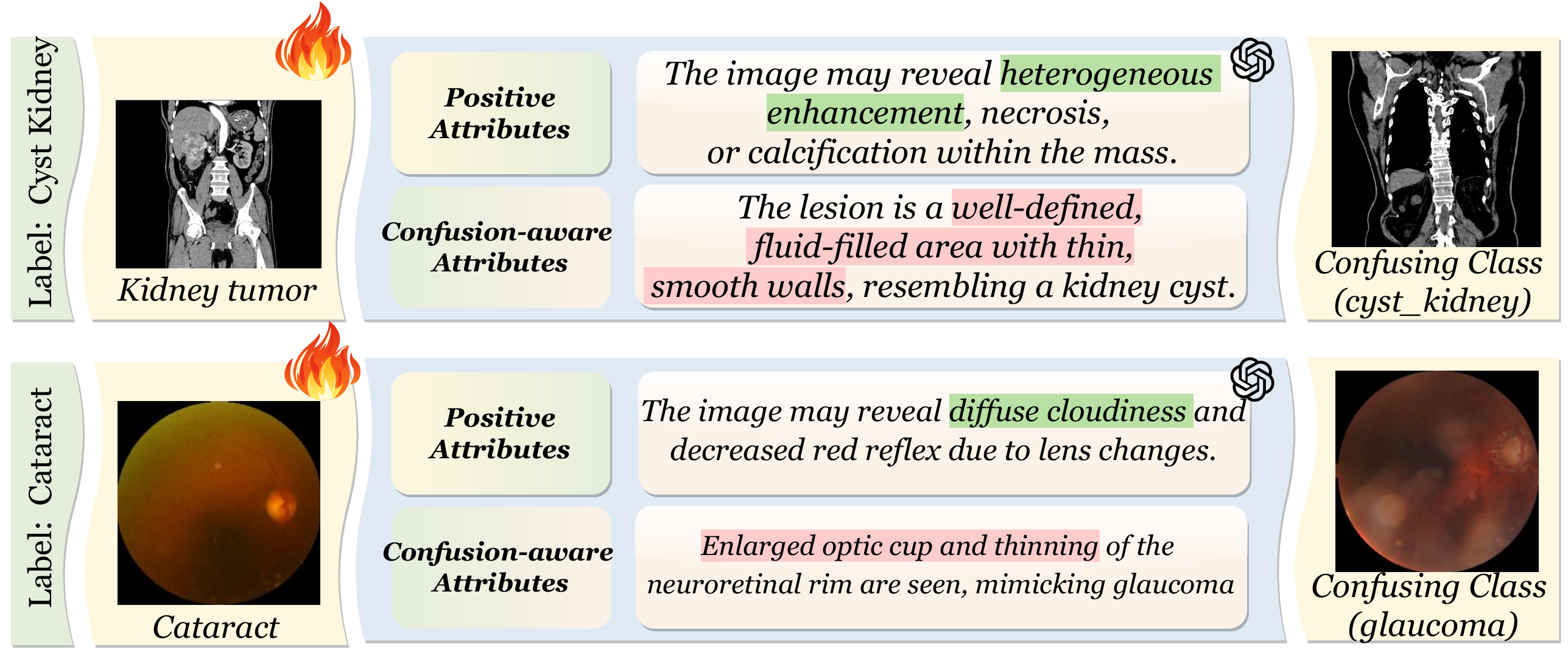}
    \caption{
{Examples of confusion-aware attributes.}
BioMedVR constructs positive attributes for discriminative cues and LLM-generated confusion-aware attributes mimicking visually similar negatives, enabling the negative expert to suppress confusion.
}
    \label{fig:confusion-examples}
\end{figure*}

\begin{table*}[ht!]
\centering
\caption{
Performance (\%) comparison of BioMedVR variants and baselines across 11 medical
datasets. \textcolor{darkgreen}{Green} indicates accuracy drops relative to the corresponding
BioMedVR backbone. Ablations (w/o CA (Confusion-aware Attributes), w/o MoPE, w/o CS Loss) show the contribution
of each component, while different LLM-generated attribute sets (GPT-4o-mini,
GPT-5-mini, GPT-5) demonstrate the robustness of BioMedVR to text--semantic sources.
Results on RN50 and ViT-B/32 further confirm consistent gains across architectures.
}

\definecolor{darkred}{RGB}{0,120,0}
\definecolor{darkgreen}{RGB}{0,120,0}
\resizebox{\textwidth}{!}{
\begin{tabular}{c|ccccccccccc|c}
\toprule
Model & BUSI & Knee X-ray & Kvasir & Lung Colon & OCTMNIST & BTMRI & CHMNIST & COVID-19 & CT-Kidney & DermaMNIST & Retina & Avg. \\
\midrule
BioMedVR-ViT-B/16 & 82.6 & 45.7 & 80.2 & 94.7 & 80.3 & 81.7 & 84.5 & 77.4 & 74.0 & 65.3 & 74.1 & 76.4 \\
w/o CA & 81.4 & 39.4 & 78.2 & 94.2 & 80.8 & 80.3 & 85.6 & 76.7 & 27.6 & 67.8 & 73.5 & 71.4 {\textcolor{darkred}{(-5.0)}} \\
w/o MoPE & 80.1 & 41.9 & 81.0 & 94.9 & 79.5 & 82.9 & 86.2 & 74.3 & 72.3 & 66.3 & 73.9 & 75.6 {\textcolor{darkred}{(-0.8)}} \\
w/o CS Loss & 80.9 & 40.9 & 76.6 & 93.4 & 75.7 & 74.7 & 85.0 & 74.6 & 72.4 & 63.1 & 72.1 & 73.6 {\textcolor{darkred}{(-2.8)}} \\

\hline
\rowcolor[rgb]{0.949,0.973,0.988}
{w. gpt-4o-mini} & 80.5 & 44.1 & 82.7 & 94.9 & 78.1 & 79.1 & 86.1 & 78.3 & 76.9 & 65.8 & 73.5 & 76.4 \\
\rowcolor[rgb]{0.949,0.973,0.988}
{w. gpt-5-mini} & 80.1 & 40.5 & 81.1 & 95.0 & 78.2 & 79.7 & 85.4 & 76.5 & 74.1 & 65.1 & 73.1 & 75.3 \\
\rowcolor[rgb]{0.949,0.973,0.988}
{w. gpt-5} & 80.1 & 40.1 & 79.9 & 94.8 & 79.8 & 80.8 & 85.3 & 77.4 & 71.6 & 63.0 & 74.0 & 75.2 \\

\hline
\rowcolor[rgb]{0.994,0.966,0.949}
BioMedVR-\textbf{RN50} & 58.5 & 38.5 & 47.4 & 77.7 & 50.3 & 58.9 & 67.0 & 54.8 & 40.7 & 36.6 & 36.9 & 51.6 \\
\rowcolor[rgb]{0.994,0.966,0.949}
Baseline (RN50) & 50.0 & 36.9 & 44.7 & 74.3 & 33.8 & 49.4 & 68.4 & 48.8 & 40.7 & 32.0 & 59.9 & 49.0 {\textcolor{darkred}{(-2.6)}} \\
\rowcolor[rgb]{0.994,0.966,0.949}
BioMedVR-\textbf{ViT-B/32} & 65.7 & 36.0 & 62.1 & 89.2 & 29.9 & 70.2 & 74.7 & 55.7 & 61.6 & 54.5 & 66.0 & 60.5 \\
\rowcolor[rgb]{0.994,0.966,0.949}
Baseline (ViT-B/32) & 61.0 & 38.2 & 57.8 & 85.6 & 44.4 & 56.3 & 75.0 & 55.9 & 59.3 & 54.0 & 61.6 & 59.0 {\textcolor{darkred}{(-1.5)}} \\
\bottomrule
\end{tabular}}
\label{exp:abl}

\end{table*}

\noindent\textbf{Visualization Examples.}
\autoref{fig:confusion-examples} shows how BioMedVR distinguishes fine-grained semantics using both positive and confusion-aware attributes.
For \textit{Kidney Tumor}, the correct class exhibits ``heterogeneous enhancement,'' while the confusing class \textit{Cyst Kidney} is characterized by a ``well-defined, fluid-filled lesion.''
By encoding cyst-like cues as confusion-aware attributes, BioMedVR learns to avoid misclassifying solid tumors as cysts.
For \textit{Cataract}, the positive attributes describe ``diffuse cloudiness,'' whereas the confusing class \textit{Glaucoma} presents cues such as ``optic cup thinning.''
This contrast allows the model to suppress misleading similarities and improve differential diagnostic accuracy.

\noindent\textbf{Visualization Results of Embedding Space.}
We visualize the t-SNE distributions of the visual embeddings on the Kvasir using the ViT-B/16.
As shown in \autoref{fig:tsne_kvasir}, the embeddings produced by AttrVR (left) exhibit significant overlap among classes such as \textit{polyps}, \textit{dyed\_resection\_margins}, and \textit{dyed\_lifted\_polyps}, indicating severe inter-class confusion.
In contrast, BioMedVR (right) yields more compact and clearly separated clusters, effectively enlarging the inter-class margins while preserving intra-class consistency.
These results demonstrate that the BioMedVR facilitates more discriminative and semantically structured feature representations in the embedding space.

\noindent\textbf{Ablation Studies.}
To verify the effectiveness of each component, we conduct a detailed ablation analysis across 11 medical datasets (\autoref{exp:abl}).
Removing the negative expert (\textit{w/o MoPE}) leads to a sharp accuracy drop of 3.1\% on COVID-19, confirming its role in confusion suppression.
Without confusion-aware attributes (\textit{w/o CA}), the negative descriptions fall back to NaN, causing ambiguous supervision and a 5\% average decline over 11 datasets, which highlights the critical role of confusion-aware signals in stabilizing training and improving discrimination.
Excluding the CS Loss (\textit{w/o CS Loss}) leads to less stable learning and higher cross-dataset variance, showing that margin calibration prevents overconfidence on ambiguous samples ({Calibration metrics and reliability diagrams are provided in the supplementary material}). Combining all components achieves the best overall performance, confirming the synergistic benefit of MoPE design.
{Beyond accuracy, we provide supplementary analyses on calibration (ECE and reliability diagrams) as well as human-vs-LLM attribute comparisons to further verify that BioMedVR is LLM-assisted rather than LLM-dependent.}

\noindent\textbf{Hyper-parameter Analyses.}
We investigate the sensitivity of BioMedVR to the CS loss weight $\beta$ and margin $m$.
As shown in \autoref{fig:params}, the model achieves stable performance across a wide range of settings,
with the optimal accuracy obtained at $\beta{=}0.3$ and $m{=}0.5$.
This shows that BioMedVR is robust to moderate variations in hyperparameters and balances discrimination and confusion suppression.

\noindent\textbf{MoE Framework Analyses.}
We compare standard \textit{MoE} architectures with our tailored \textit{Confusion-aware MoPE}.
As shown in \autoref{fig:params}, increasing the number of generic MoE experts yields limited gains,
while the 2-MoPE---comprising experts specialized for positive, and confusion-aware attributes---achieves the best accuracy (82.6\%).
This demonstrates that semantic specialization, rather than expert quantity, is key to effective VR.
Further details on MoE can be found in the supplementary material.

\noindent\textbf{Effect of LLM-Generated Confusion-aware
Attributes.}
We analyze the influence of different LLMs used for generating attribute descriptions.
As shown in \autoref{exp:abl}, BioMedVR achieves the highest average accuracy (76.4\%) when using GPT-4.1-generated attributes.
Alternative models (GPT-4o-mini, GPT-5-mini, GPT-5) yield slightly lower but comparable results \textcolor{red}{(0\% to -1.2\%)},
indicating that BioMedVR is robust to the choice of LLM and benefits from the richer semantics captured by larger models.

We further compare the performance of BioMedVR using human-generated vs.\ LLM-generated attributes in the supplementary material.

\section{Conclusion}
In this work, we present BioMedVR, a confusion-aware MoPE framework that brings VR to biomedical imaging for the first time. Unlike prompt learning, which depends on model internals and struggles with fine-grained medical ambiguity, BioMedVR provides an input-space, architecture-agnostic, and privacy-preserving adaptation mechanism for VLMs.
By decoupling visual prompts into a positive expert for discriminative alignment and a negative expert guided by confusion-aware attributes, together with a confusion-suppression loss, BioMedVR effectively mitigates inter-class confusion among visually similar diseases. Experiments on 11 biomedical datasets and 7 natural-image benchmarks show that BioMedVR achieves SOTA few-shot and zero-shot performance, consistently surpassing prior VR and prompt-learning methods, with visualizations confirming clearer embedding separation. We believe this work provides a new paradigm for efficient and interpretable medical adaptation of VLMs, paving the way toward reliable and data-efficient AI-assisted diagnosis in healthcare.

\section*{Limitations.}

BioMedVR is designed for confusion-aware recognition and differential-diagnosis-inspired discrimination rather than full clinical decision support.
The confusion-aware attributes are generated offline and may depend on the quality of the underlying language model and prompts.
Future work will incorporate expert-curated candidate lists and clinically grounded evaluation protocols to further validate real-world diagnostic utility.

\bibliographystyle{splncs04}
\bibliography{main}
\end{document}